\RequirePackage{fix-cm}
\documentclass[twocolumn]{svjour3}          
\smartqed  
\usepackage{graphicx}
\usepackage{amsmath,amssymb,amsfonts}
\usepackage{algorithmic}
\usepackage{textcomp}
\usepackage{url}
\usepackage{float}
\usepackage{tabularx}
\usepackage{upquote}
\usepackage{lipsum}
\usepackage{color}
\usepackage{hhline}
\usepackage{mathtools}
\usepackage[numbers]{natbib}

\begin{document}

\title{Unsupervised Deep Context Prediction for Background Estimation and Foreground Segmentation
}


\author{Maryam Sultana        \and
            Arif Mahmood \and
            Sajid Javed \and 
            Soon Ki Jung           
}


\institute{Maryam Sultana, Soon Ki Jung \at
              School of Computer Science and Engineering \\
              Virtual Reality Laboratory,\\
              Kyungpook National University,\\
              Daegu, Republic of Korea,\\
              \email{maryam@vr.knu.ac.kr, skjung@knu.ac.kr}           
           \and
           Arif Mahmood \at
           Department of Computer Science and Engineering \\
           Qatar University\\
           Doha, Qatar \\
  \email{rfmahmood@gmail.com}
          \and
          Sajid Javed \at
          Department of Computer Science\\
          Tissue Image Analytics Laboratory\\
          University of Warwick\\
          United Kingdom,\\
      \email{s.javed.1@warwick.ac.uk}
}

\date{Received: date / Accepted: date}

\maketitle

\begin{abstract}
In many high level vision applications such as tracking and surveillance, background estimation is a fundamental step. In the past,  background estimation was usually based on low level  hand-crafted features such as raw color components, gradients, or local binary patterns. These existing algorithms  observe performance degradation in the presence of various challenges such as dynamic backgrounds, photo-metric variations, camera jitter, and shadows. To handle these challenges for the purpose of accurate background estimation, we propose a unified method based on Generative Adversarial Network (GAN) and image inpainting. It is an unsupervised visual feature learning hybrid GAN based on context prediction. It is followed by a semantic inpainting network  for texture optimization. We also propose a solution of arbitrary region inpainting by using  center region inpainting and  Poisson blending. The proposed  algorithm is compared with the existing algorithms for background estimation on SBM.net dataset and for foreground segmentation on CDnet 2014 dataset. The proposed  algorithm has outperformed the compared methods with  significant margin.  


\end{abstract}
\section{Introduction}
Background estimation and foreground segmentation is a fundamental step in several  computer vision applications, such as salient motion detection \cite{2PRPCA}, video surveillance \cite{bouwmans2014robust},  visual object tracking \cite{zhang2015robust} and moving objects detection \cite{viola2001rapid, ren2015faster, girshick2016region}. The goal of background modeling is to efficiently and accurately extract a model which describes the scene in the absence of any foreground objects. 
Background modeling becomes challenging in the presence of dynamic backgrounds, sudden illumination variations, and camera jitter which is mainly induced by the sensor. 
A number of techniques have been proposed in the literature that mostly address relatively simple scenarios for scene background modeling \cite{bouwmans2017scene}, because complex background modeling is a challenging task itself specifically in handling real-time environments.\\
\begin{figure}[t]
	\centering
	\includegraphics[scale=0.34]{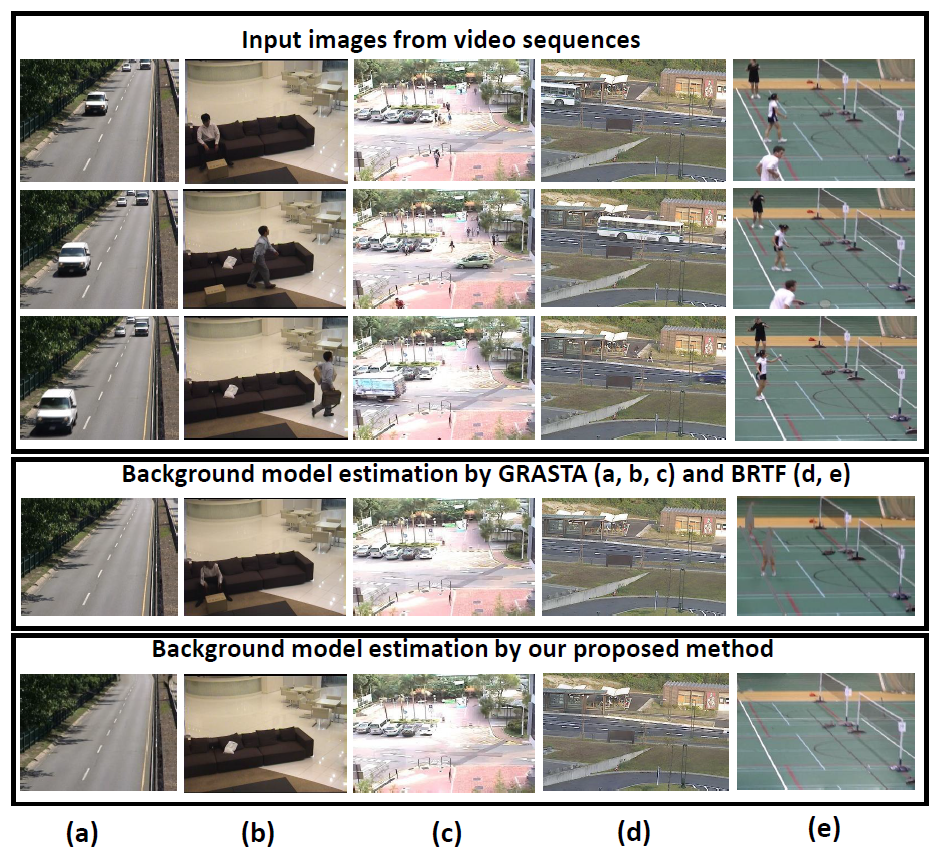}
	\caption{Estimated background images from the SBM.net dataset : Sequences in (a) are from the category "Basic" named "Highway". (b) Sequence "Sofa" from the category "Intermittent Object Motion" (c) Sequence "Chuk Square" from the category "Very Short". (d) Sequence "Bus Station" from the category "Very Long". (e) Sequence "Badminton" is also from the category "Jitter". In almost all of these cases, for accurate the background estimation, the average gray-level error (AGE) is less in our proposed algorithm as mentioned in Table \ref{all_dataset}.}
	\label{fig_intro}
\end{figure}
To solve the problem of background subtraction, Stauffer \textit{et al.} \cite{stauffer1999adaptive} and Elgammal \textit{et al.} \cite{elgammal2000non} presented methods based on statistical background modeling. It starts from an unreliable background model which identify and correct initial errors during the background updating stage by the analysis of the extracted foreground objects from the video sequences. Other methods proposed over the past few years also solved background initialization  as an optimal labeling problem \cite{nakashima2011automatic, park2013unified, xu2008loopy}. These methods compute label for each image region, provide the number of the best bootstrap sequence frame such that the region contains background scene. Taking into account spatio-temporal information, the best frame is selected by minimizing a cost function. The background information contained in the selected frames for each region is then combined to generate the background model.
The background model initialization methods based on missing data reconstruction  have also been proposed \cite{sobral2015comparison}. These methods work where missing data are due to foreground objects that occlude the bootstrap sequence. Thus, robust matrix and tensor completion algorithms \cite{sobral2017matrix} as well as inpainting methods \cite{colombari2005exemplar} have shown to be suitable for background initialization. More recently, deep neural networks are introduced for image inpainting \cite{pathak2016context}. In particular, Chao Yang et al. \cite{yang2016high} used a trained CNN (Context Encoder \cite{pathak2016context}) with combined reconstruction loss and adversarial loss \cite{goodfellow2014generative} to directly estimate missing image regions. Then a joint optimization framework updates the estimated inpainted region with fine texture details. This is done by hallucinating the missing image regions via modeling two kinds of constraints, the global context based and the local texture based, with convolutional neural networks. This framework is able to estimate missing image structure, and is very fast to evaluate. Although the results are encouraging but it is unable to handle random region inpainting task with fine details.

\begin{figure*}[t]
	\centering
	\includegraphics[scale=0.42]{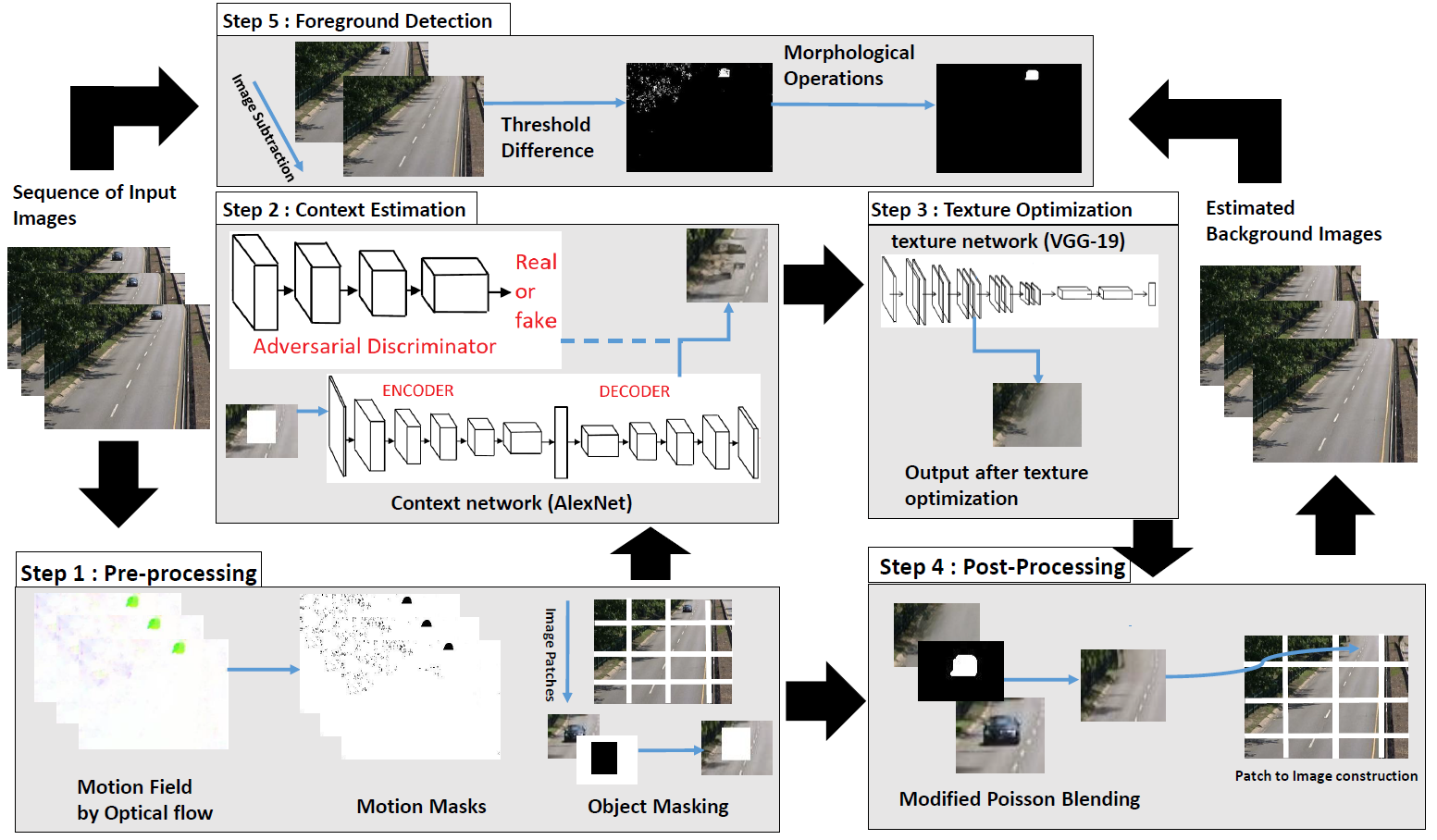}
	\caption{Workflow of the proposed algorithm for background estimation. Step (1) describes the motion
		estimation via dense optical flow, the creation of motion masks, image to patch conversion and object masking for center region inpainting task. Step (2) evaluates the prediction of missing region with context prediction hybrid GAN network. In step (3), to improve the fine texture details of the predicted context, the output of step (2) is given to texture network. Step (4) In the previous step it can be seen that in this case some information of road (white lines in the middle) is being missed by texture network so, Modified Poisson Blending technique is applied to get final results. Step (5) we threshold the difference of the estimated background via DCP and current frame of the video sequence. Afterwards the thresholded difference is binarized and run through extensive Morphological Operations to extract the foreground moving object.}
	\label{fig_frame}
\end{figure*}

In this paper we propose to predict missing image structure using inpainting method, for the purpose of scene background initialization. We name our method as \textit{Deep Context Prediction} (DCP), because it has the ability to predict context of a missing region via deep neural networks. Few visual results of the proposed DCP algorithm are shown in  Figure \ref{fig_intro}. 
Given an image, fast moving foreground objects are removed  using motion information leaving behind missing image regions (see Figure \ref{fig_frame} Step (1)). We train a convolutional neural network to estimate  the missing pixel values via inpainting method. The CNN model consists of an encoder capturing the context of the whole image into a latent feature representation and a decoder which uses this representation to produce the missing content of the image. The model is closely related to auto-encoders \cite{bengio2009learning, hinton2006reducing}, as it shares a similar architecture of encoder-decoder. Our contributions in the proposed method are summarized as follows:
\begin{itemize}
	
	\item We extract the temporal information in the video frames by using dense optical flow \cite{liu2009beyond}. After mapping motion information to motion mask, we are able to approximately identify fast moving foreground objects. We eliminate these objects and fill the missing region using the proposed DCP algorithm by estimating background.

	\item In our proposed DCP method, we  train a context encoder similar to \cite{yang2016high} on scene-specific data. The network is pre-trained on ImageNet dataset \cite{imagenet_cvpr09}. DCP is a joint optimization framework that can estimate context of missing regions by inpainting in central shape and later transform this predicted information to random regions by the help of Modified Poisson Blending (MPB) \cite{afifi2015mpb} technique. The framework is based on two constraints, a global context based which is a hybrid GAN model trained on scene-specific data and a local texture based which is VGG-19 network \cite{simonyan2014very}.
	
	
	\item For the purpose of foreground object detection, we first estimate background  via DCP, and later we binarize the difference of the background with the current frame, leading to more precise detection of foreground moving objects. This binarized difference is enhanced through morphological operations to remove false detection and noisy pixel values.
\end{itemize}

The proposed DCP algorithm is based on context prediction,  therefore it can predict homogeneous or blurry contexts more accurately compared to other background initialization algorithms. 
In case of background motion, DCP can still estimate background  by calculating motion masks via optical flow, as our target is to eliminate foreground moving objects only. DCP is also not effected by intermittent object motion because of the same reason mentioned previously. In challenging weather conditions (rain, snow, fog) dense optical flow can identify foreground moving objects, so targeting only those objects to remove and inpaint them with background pixels makes DCP a good background estimator. For the case of difficult light conditions DCP can estimate background accurately because of homogeneity in the context of scenes with low illumination.
\section{Related Work}
Over the past few years, background subtraction and foreground detection has remained the part of many key research studies \cite{cao2016total, ortego2016rejection, haines2014background, javed2016motion, javed2015robust} as well as scene background initialization \cite{bouwmans2017scene, bouwmans2014robust, erichson2016randomized, maddalena2015towards, ye2015foreground}. In the problem of background subtraction, the critical step is to improve the accuracy of the detection of foreground. On the other hand, the task of estimating
an image without any foreground is called scene background modeling. Many comprehensive studies have  been conducted to this problem \cite{bouwmans2017scene, bouwmans2014robust, maddalena2015towards}. Gaussian Mixture Model (GMM) \cite{stauffer1999adaptive, zhang2013mining, zivkovic2004improved, varadarajan2013spatial, lu2014multiscale} is a well known technique for background modeling. It uses probability density functions as mixture of Gaussians  to model color intensity variations at pixel level. Recent advances in GMM  include minimum spanning tree \cite{chen2017spatiotemporal} and bidirectional analysis \cite{shimada2013background}. On the other hand most GMM based methods also suffer performance degradation in complex and dynamic scenes.

In the past, particularly for the problem of background modeling many research studies have been conducted by using \textit{Robust Principal Component Analysis} (RPCA). Wright \textit{et al.} \cite{wright2009robust} presented
the first proposal of RPCA-based method which has the ability to handle the outliers in the input data. Later Candes \textit{et al.} \cite{candes2011robust} used RPCA for background modeling and foreground detection. Beyond good performance, RPCA-based methods are not ideal for real-time applications because these techniques possess high computational complexity. Moreover, conventional RPCA-based methods process data in batch manner. Batch methods are not suitable for real-time applications and  mostly work offline. Some online and hybrid RPCA based methods have also been presented in the literature to handle the batch problem \cite{javed2017background} while global optimization is still a challenge in these approaches \cite{javed2015robust, he2012incremental, xu2013gosus}.
Xiaowei Zhou \textit{et al.} \cite{DECOLOR} proposed an interesting technique known as \textit{Detecting Contiguous Outliers in the LOw-rank Representation} (DECOLOR). Limitation of no prior knowledge in RPCA based methods on the spatial distribution of outliers leads to develop this technique. Outliers information is modeled in this formulation by using Markov Random Fields (MRFs). 
Another online RPCA algorithm proposed by Jun He \textit{et al.} \cite{he2012incremental} is \textit{Grassmannian Robust Adaptive Subspace Tracking Algorithm} (GRASTA). It is an online robust subspace tracking algorithm embedded with traditional RPCA. This algorithm operates on data which is highly sub-sampled. If the observed data matrix is corrupted by outliers as in most cases of real-time applications, $l^{2}$-norm based objective function is  best-fit to the subspace.
Hybrid Approach: use a time window to obtain sufficient context information then process it like a small batch.
Recently S. Javed \textit{et al.} \cite{MSCL} proposed a hybrid  technique named \textit{Motion-assisted Spatiotemporal Clustering of Low-rank} (MSCL) based on RPCA approach.
In this method for each data matrix, sparse coding is applied
and estimation of the geodesic subspace based Laplacian matrix is calculated. The normalized Laplacian matrices estimated over both distances Euclidean as well as Geodesic are embedded into the basic RPCA framework.
In 2015 Liu \textit{et al.} \cite{godec} developed a technique called \textit{Sparse Matrix Decomposition} (SSGoDec), which is capable of efficiently and robustly estimating the low-rank part $L$ of background and the sparse part S of an input data matrix $D = L + S + G$ with a factor of noise $G$. This technique alternatively assigns the low-rank approximation of difference between input data matrix and sparse matrix $(D-S)$ to $L$. Similarly it also assigns the vice verse as well which is the sparse approximation of $(D-L)$ to $S$. To overcome the batch constraint of RPCA based methods J. Xu \textit{et al.} \cite{gosus} presented a method called \textit{Grassmannian Online Subspace Updates with Structured-sparsity} (GOSUS). Although this method performs well for background estimation problem but global optimality is still the challenging issue in this approach. 
Qibin Zhao \textit{et al.} \cite{zhao2016bayesian} presented a method called \textit{Bayesian Robust Tensor Factorization for
	Incomplete Multiway Data} (BRTF). This method is a generative model for robust tensor factorization in the presence of missing data and outliers.
X. Guo \textit{et al.} \cite{guo2014robust} presented a method called \textit{Robust Foreground Detection Using Smoothness and Arbitrariness Constraints} (RFSA). In this method the authors considered the smoothness and the arbitrariness of static background, thus formulating  the problem in a unified framework from a probabilistic perspective. \\
Recently, Convolutional Neural Network (CNN) based methods have also shown significant performance for foreground detection by scene background modeling \cite{braham2016deep, wang2017interactive, zhang2015deep}. For instance, Wang \textit{et al.} \cite{wang2017interactive} proposed a simple yet effective supervised CNN based method for detecting moving objects in static background scenes. CNN based methods perform best in many complex scenes however, our proposed method DCP is unsupervised therefore it do not require any labelled data for training purposes.

\section{Proposed Method} \label{Proposed} 
Our proposed background foreground separation technique has five steps. 1.) Motion masks evaluation via dense optical flow. 2.) Estimation of missing background pixels by Context Encoder (CE). 3.) The improvement of estimated missing pixels texture by a multi-scale neural patch synthesis. 4.) Modified Poisson Blending technique is applied to get final results. 5.) The foreground objects are detected by applying threshold on the difference between the estimated background from DCP and the current frame, which is later enhanced by morphological operations. The work flow diagram of DCP is shown in  Figure \ref{fig_frame}. Detail description of the above mentioned steps is as follows:

\subsection{Motion Masks via Optical Flow}
For the purpose of background estimation from the video frames, we have to first identify the fast moving foreground objects. These objects are recognized by using optical flow \cite{liu2009beyond} which is then used to create a motion mask. 
Dense optical flow is calculated between each pair of consecutive frames in the given input video sequence $S$. Motion mask $M$ is computed by using motion information from a sequence of video frames. Let $S_{t}$ and $S_{t-1}$ be the two consecutive frames in $S$ at time instant $t$ and $t-1$, respectively. Considering $v^{y}_{t,p}$ be the vertical component and  $u^{x}_{t,p}$ be the horizontal component of the motion vector at position $p$ which is computed between consecutive frames. The corresponding motion mask, $m_{t}\in\{0,1\}$ will be computed as :
\begin{equation}
m_{t,p}=\begin{cases}
1,~~if~~\sqrt[]{(u^{x}_{t,p})^2 + (v^{y}_{t,p})^2 }< t_h,\\
0,~~otherwise. 
\end{cases}
\end{equation}
In the above equation, $t_h$ is threshold of motion magnitude. It is computed by taking the average of all pixels in the motion field. Selection of the threshold $t_h$ is adapted in such a way that all pixels in $S$ consisting of  motion greater than $t_h$  belongs to the foreground. In order to avoid noise in the background, threshold $t_h$ is selected to be large enough.

\subsection{Background Pixels Estimation via Context Prediction} \label{CE}
Given image patches from a video with missing regions such as foreground object regions, we predict context via context encoder \cite{pathak2016context}. The context encoder is a hybrid GAN model which is trained on the basis of convolutional neural network to estimate the missing pixel values. It consists of two parts: an encoder which captures the context of a given image patch into a compact latent feature space. While the other part is a decoder which uses encoded representation to produce missing image patch content. Overall architecture of context encoder is a simple encoder-decoder pipeline. 

The encoder is derived from the AlexNet \cite{krizhevsky2012imagenet}, however the network is not trained for classification, rather it is trained for context prediction. Training is performed on ImageNet as well as using scene-specific video sequences patches. 
In order to learn an initial context prediction network, we train a regression network $F$ to get  response $F(x_{m})$, where $x_m$ is  the input image patch $x$ pixel-wise multiplied by object mask $m_{o}$: $x_m=x \odot m_o$. Since $m_o$ is a binary mask with fix central region which covers the whole object in motion mask.

The patch $x_m$ with a missing region $H$, is input to Context Network. The response $F(x_{m})$ of trained context network is estimated via joint loss functions to estimate the background $x_b$ in the missing region $H$. We have experimented with two joint loss functions including reconstruction loss $L_{rec}$ and adversarial loss $L_{adv}$ \cite{pathak2016context}. The reconstruction loss $L_{rec}$ is defined as:
\begin{equation}
L_{rec}(x_{m},x_b,H)=||F(x_{m}) - C(x_b, H)||_{2}^{2}.
\end{equation}
The adversarial loss is given as:
\begin{equation}
\begin{split}
L_{adv}(x_{m},x_b,H)= \max\limits_{D}~E_{x_{m}\in\chi_{m}}[log(D(C(x_b,H)))\\
+log(1-D(F(x_{m})))],
\end{split}
\end{equation}
where $D$ is the adversarial discriminator and $C(\cdot)$ defines the
operation of extracting a sub-image in the central region during inpainting process. Overall loss function is a linear combination of both reconstruction and adversarial losses. 
\begin{equation} \label{eta_eq}
L=\eta L_{rec}(x_{m},x_b,H) + (1-\eta)L_{adv}(x_{m},x_b,H),   
\end{equation}
where $\eta$ is a relative weight of each loss function. 

\subsection{Texture Optimization of Estimated Background} \label{texture}
In the last section, we estimated a background patch $x_b$ via Context Encoder (CE). But the estimated context still contains irregularities and blurry texture at low resolution of the image patch. To solve this blurry estimated context problem for high resolution inpainting with fine details, we use texture network at three-level pyramid of image patches. This network optimizes over three loss terms: the predicted context term initialized by CE, the local texture optimization term, and the gradient loss term. The context prediction term captures the semantics including global structures of the image patches. The texture term maps the local statistics of input image patch texture, and the gradient loss term enforces the smoothness between the estimated context and the original context. For three-level pyramid approach the test image patch is assumed to be always cropped to $512\times512$ with a $256\times256$ hole in the center at fine level. However with step-size two, downsizing to the coarse level as $128\times128$ size image patch with a $64\times 64$ missing region is initialized by CE. Afterwards context of missing region is estimated in a coarse-to-fine manner. At each scale, the joint optimization is performed to update the missing region and then upsampling is done to initialize the joint optimization which sets the context constraint for the next scale of image patch. This process repeats this until the joint optimization is completed at the fine level of the pyramid. The texture optimization term is computed using the $VGG-19$ \cite{simonyan2014very} which is pre-trained on ImageNet.

Once the context is initialized by CE at the coarse scale, we use the output $F(x_{m})$ and the original image as the initial context constraint for joint optimization. Let $x_o$ be the original image patch with missing region  filled with the CE. Upsampled version of $x_o$ are used as the initialization for joint optimization at the fine scales.


For the input image patch $x_{o}$ we would like to estimate the fine texture of the missing region. 
The region corresponding to $x_o$ in the feature map of $VGG-19$ network is $\psi(x_o)$  and $\psi(H)$ is the feature map corresponding to the missing region. For texture optimization $C(\cdot)$ also defines the operation of extracting a sub-feature-map in a rectangular region, i.e. the context of $\psi(x_o)$ within $\psi(H)$ is returned by $C(\psi(x_o), H)$.

The optimal solution for accurate reconstruction of the missing content is obtained by minimizing the following objective function at each scale $i=1,2...,n$. 
\begin{equation}\label{objectivefunction}
\begin{split}
\hat{x}^{i+1} = arg~min~E_{CE}(C(x_o, H),C(x_{o}^{i}, H)) \\
+\gamma E_{T}(\psi_{T}(x_o), \psi(H) + \delta \Pi(x_o),
\end{split}
\end{equation}
where $C(x_{o}^{1}, H) = F(x_{o}),~\psi_{T}(.)$ represents a feature maps in the texture network $T$ at an intermediate layer, $\gamma$ and $\delta$ are weighting reflecting parameters. \cite{yang2016high}. The first term $E_{CE}$ in equation \eqref{objectivefunction} is context constraint which is defined by the difference between the previous context prediction and the optimization result:
\begin{equation}
E_{CE}(C(x_o, H), C(x_o^{i}, H)) = || C(x_o, H) - C(x_o^{i}, H) ||^{2}_{2}.
\end{equation}
The second term $E_{T}$ in equation \eqref{objectivefunction} handles the local texture constraint, which minimizes the inconsistency of the texture appearance outside and inside the missing region. We first select a single feature layer or a combination of different feature layers in the texture network $T$, and then extract its feature map $\psi_{T}$. In order to do texture optimization, for each query local patch $P$ of size $w\times w\times c$ in the missing region $\psi(H)$, our target is to find the most similar patch outside the missing region, and calculate  loss as mean of  the query local patch and its nearest neighbor distances.
\begin{equation}
\begin{split}
E_{T}(\psi_{T}(x_o), H) =  \\ 
\frac{1}{|\psi(H)|} \sum \limits_{i\in \psi(H)} || C(\psi_{T}(x_o), P_{i}) - C(\psi_{T}(x_o), P_{np(i)}) ||^{2}_{2},
\end{split}
\end{equation}
In the above equation, the local neural patch centered at location $i$ is $P_{i}$, the number of patches sampled in the region $\psi(H)$ is given by $|\psi(H)|$, and $np(i)$ is the calculated as:
\begin{equation}
np_{i} = \arg\min\limits_{j\in n(i)\wedge j\notin H^{\psi}}|| C(\psi_{T}(x), P_{i}) - C(\psi_{T}(x), P_{j}) ||^{2}_{2},  
\end{equation}
where $n(i)$ is the set of neighboring locations of $i$ excluding the overlap with the missing unknown region $\psi(H)$.
We also add the gradient loss term to encourage smoothness in texture optimization \cite{yang2016high}:
\begin{equation}
\begin{split}
\Pi(x_o) = \sum\limits_{j,k} ((x_o(j,k+1) - x_o(j,k))^2 + \\
(x_o(j+1,k) - x_o(j,k))^2),    
\end{split}
\end{equation}

\subsection{Blending of Estimated and Original Textures}
After the texture optimization, some information around the central region during inpainting process is being missed or removed due to rectangular shaped region as shown in figure \ref{fig_frame}. In order to change the rectangular shaped predicted context to the irregular shaped region, Modified Poisson Blending technique (MPB) \cite{afifi2015mpb} is used. It is based on Poisson image editing for the purpose of seamless cloning. The MPB technique has three steps, the first step, uses the source image which is inpainted image via DCP as a known region and the target image which is original image containing foreground as an unknown region. Afterwards it requires motion mask by optical flow around the interested object in the source image for solving Poisson
equation \cite{perez2003poisson} under gradient field and predefined
boundary condition. MPB technique has few  modifications to Poisson image editing technique that eliminates the bleeding problems in the composite image by using Poisson blending with fair
dependency of source which is inpainted context and target pixels which are original image pixels. 
In the next step, MPB technique uses the composite image as unknown region
and the target image with foreground object as a known region. After applying Poisson blending algorithm, we get another composite image which will be used in third step. To reduce bleeding artifacts, MPB technique generate an alpha mask that is used to combine both composite images from previous steps to get final image that is free from color bleeding. In practice 
this method helps in discarding the useless information which came along rectangular region inpainting
process.
\begin{figure*}[t]
	\centering
	\includegraphics[scale=0.4]{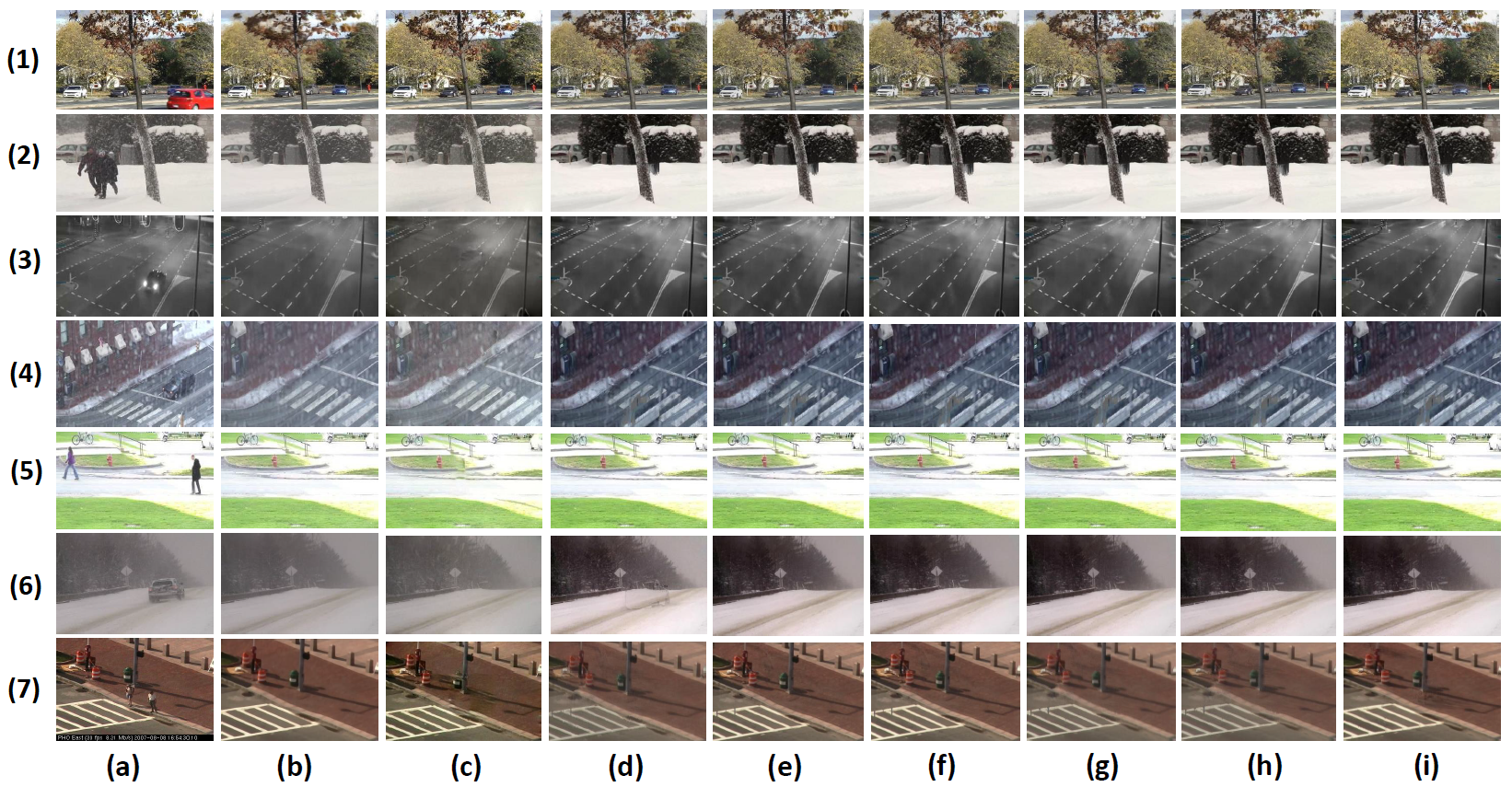}
	\caption{Qualitative results of the proposed method. (a) 7 images from the input video sequences, (b) Ground truth, (c) estimated background model by the proposed DCP method, (d) RFSA, (e) GRASTA, (f) BRTF, (g) GOSUS, (h) SSGoDec, (i) DECOLOR. From top to bottom: each input sequence is selected from different categories. (1)sequence "Fall" from "Background Motion", (2) "Skating" from "Basic", (3)"StreetCornerAtNight" from "Basic", (4) "WetSnow" from "Basic", (5) "Pedestrians" from "Very Short", (6) "Snowfall" from "Very Short", and (7) "SideWalk" from "Jitter".}
	\label{fig_vis}
\end{figure*}

\subsection{Foreground Detection}
In this work we mainly focus on the problem of background initialization. However, in this section we extend our work to foreground detection as well. Thus we are able to compare our work  with foreground detection algorithms as well, in addition to the work on background initialization only.  For the purpose of foreground detection, we threshold the difference of the estimated background via DCP and the current frame of the video sequence. The  difference is threshold and  
binarized and processed through  Morphological Operations (MO)  with suitable Structuring Elements (SE). Thus the work done in this section may be considered as post processing.

These operations first include opening operation of an image which is erosion followed by the dilation with the same SE: 
\begin{equation}\label{open}
    I \circ SE = (I \ominus SE) \oplus SE,
\end{equation}
where $I$ is the binarized  difference, $\ominus$ and $\oplus$ denote erosion and dilation respectively. Afterwards, closing operation is performed on  this  image. It is in reverse way, that is dilation followed by erosion with same SE, but different from SE used in the opening operation. \begin{equation}
 I' \bullet SE = (I' \oplus SE) \ominus SE.  
\end{equation} 
Now here $I'$ is the difference image from equation \eqref{open}, $\oplus$ and $\ominus$ denote dilation and erosion respectively. Successive opening and closing of the binarized difference frame with proper SE leads us to separate the foreground objects from the background. The choice of the SE is very crucial in successive opening and closing of the binarized difference frame as it may lead to false detection if not selected according to the shape of the objects in the video frames. 
The MO not only fills the missing regions in thresholded difference but also removes the unconnected pixels values of background which are considered to be noise in foreground detection process.

\section{Experiments}

Our background estimation and foreground detection techniques are based on inpainting model  similar to \cite{yang2016high}. We trained the context prediction model additionally with scene-specific data in terms of patches of size $128\times128$ for $3$ epoches. The texture optimization is done with $VGG-19$ network pre-trained on ImageNet for classification. The frame selection for inpainting the background is done by summation of pixel values in the forward frame difference technique. If the sum of difference pixels is small, then current frame is selected.

We evaluate our proposed approach on two difference datasets, including Scene Background Modeling   (SBM.net)\footnote{\url{http://scenebackgroundmodeling.net/}} for background estimation and Change Detection 2014 Dataset (CDnet2014)\cite{wang2014cdnet} for foreground detection. 
On both datasets our proposed algorithm has outperformed existing state of the art algorithms with a significant margin.

\subsection{Evaluation of Deep Context Prediction (DCP) for Background Estimation} \label{sbmdataset}
\begin{figure}[t]
	\includegraphics[scale=0.28]{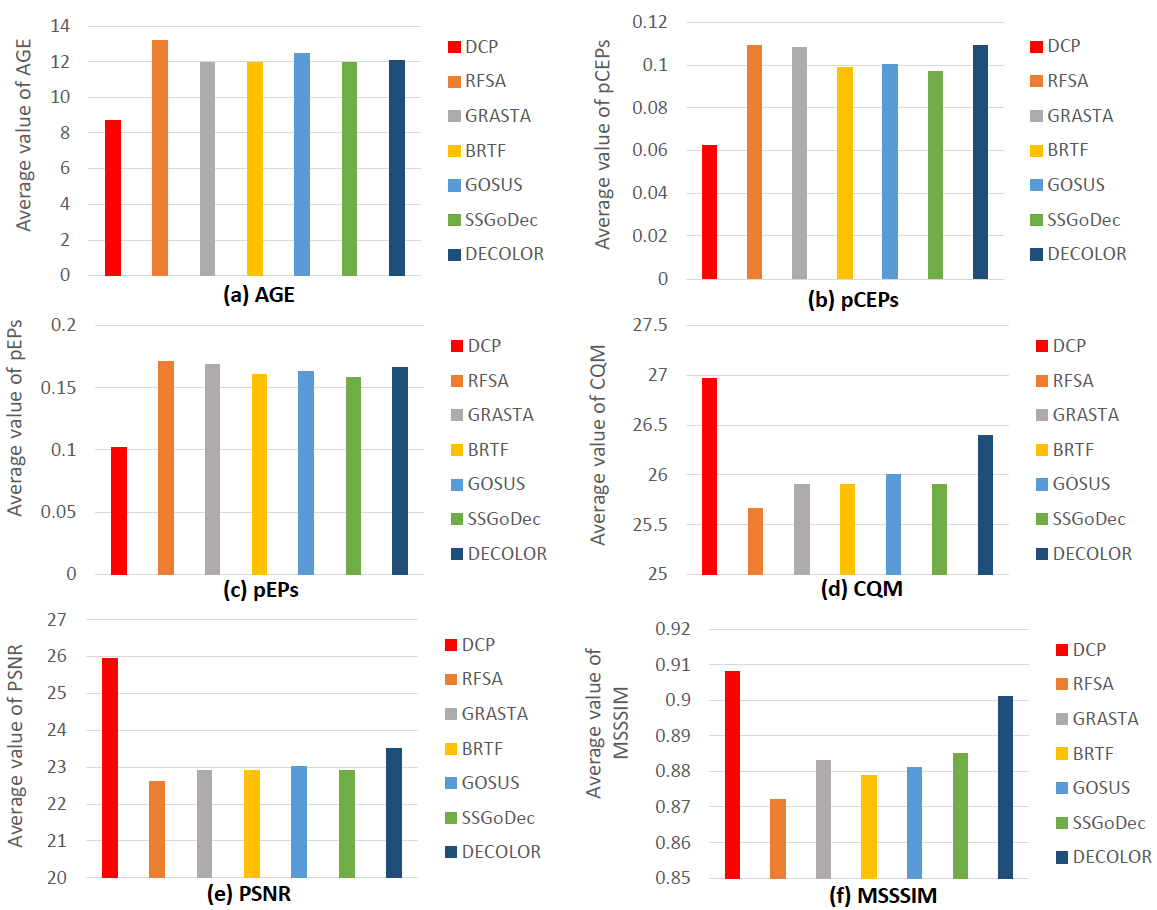}
	\caption{Average performance comparison of DCP on each metric with $6$ state-of-the-art methods on the $7$ categories of SBM.net dataset. (a) AGE (b) pCEPs and (c) pEPs, (minimum  is best). (d) CQM (e) PSNR (f) MSSSIM, (maximum  is best).}
	\label{fig_metric}
\end{figure}

We have selected all videos out of $7$ categories from SBM.net dataset as shown in  Table \ref{all_dataset}. Every category in SBM.net dataset has challenging video sequences for background modeling. 
In this experiment, results are compared with $6$ state-of-the-art methods, including RFSA \cite{guo2014robust}, GRASTA \cite{he2012incremental}, BRTF \cite{zhao2016bayesian}, GOSUS \cite{gosus}, SSGoDec \cite{godec}, and DECOLOR \cite{DECOLOR}  using  implementations of the original authors. Background estimation models are compared using Average Gray-level Error (AGE), percentage of Error Pixels (pEPs), Percentage of Clustered Error Pixels (pCEPs), Multi Scale Structural Similarity Index (MSSSIM), Color image Quality Measure (CQM), and Peak-Signal-to-Noise-Ratio (PSNR) \cite{maddalena2015towards}. For best performance the aim is to minimize AGE, pEPs, and pCEPs while maximizing MSSSIM, PSNR, and CQM (Fig. \ref{fig_metric}). The detail description of results with respect to each category is as follows:

\textbf{Category: Background Motion} contains $6$ video sequences. In this category the proposed DCP algorithm achieved  best performance among all the compared methods. The performance of DECOLOR, SSGoDec, RFSA, GRASTA and BRTF has remained quite similar with minimal difference in AGE as shown in table \ref{all_dataset}. GOSUS has the highest average gray level error among all the compared methods. Targeting only foreground objects to be eliminated and filled with background pixel values via inpainting method makes DCP to perform better in this category as compared to all other methods. The visual results are shown in  Figure \ref{fig_vis}, 1st row.
\begin{table*}[htbp]
	\caption{The AGE scores over SBM.net dataset for the six state of the art methods compared with DCP for background subtraction. The best AGE score for each video sequence is shown in blue, and the best average AGE scores for each category is shown in red.}
	\scalebox{0.75}{
		\label{all_dataset}
		\begin{tabular}{|l|l|l|l|l|l|l|l|l|}
	\hline
			Category          & Videos   & AGE \\ 
			\hline
			&           & DCP & RFSA \cite{guo2014robust}& GRASTA \cite{he2012incremental} & BRTF \cite{zhao2016bayesian} & GOSUS \cite{gosus} & SSGoDec \cite{godec} & DECOLOR \cite{DECOLOR} \\
			\hline
			Background Motion & Canoe:   & \textcolor{blue}{6.3250}	& 14.8805 & 14.9438 & 14.8798 & 14.9677 & 14.9464 & 13.6732  \\
			& Advertisement Board: & \textcolor{blue}{2.3378} & 3.4762 & 3.4812 & 3.4640 &3.4733 & 3.4742 & 3.6604   \\
			& Fall:    & \textcolor{blue}{19.0737} & 24.3364 & 24.6026 & 24.4283 &  24.5935 & 24.5702 & 24.8117  \\
			& Fountain 01: & 9.6775 & \textcolor{blue}{5.7150} & 5.7539 & 5.7383 & 5.7750 & 5.7442 & 6.2959  \\
			& Fountain 02:   & 14.0579 & 7.3288 & 7.0811 & 7.3307 & 7.0867 & 7.0801 & \textcolor{blue}{6.4137}\\
			& Overpass:   & \textcolor{blue}{6.4089} & 14.7162 & 14.7489 & 14.7183 & 14.7614 & 	14.7369 & 8.6909   \\
			\hline
			Average AGE:    &          & \textcolor{red}{9.6468}  & 11.7422 & 11.7686 & 11.7599 & 12.1183 & 11.5934 & 11.6340  \\
			\hline
			Basic:            & 511:     & \textcolor{blue}{3.5786} & 5.0972 & 6.1220 & 4.9151 & 5.2681 & 6.6025 & 7.5225 \\
			& Blurred:   & \textcolor{blue}{2.1041} & 4.9735 & 47.3112 & 4.9527 & 105.1528 & 51.9253 & 4.7345 \\
			& Camouflage FG Objects:   & \textcolor{blue}{2.5789} & 4.7951 & 5.0457 & 4.7411 &  4.3364 & 5.9418 & 4.4703  \\
			& Complex Background :   & 6.3453 & 6.8593 & 6.2202 & 6.8868 & 6.1947 & 6.1828 & \textcolor{blue}{5.6215}   \\
			& Hybrid :   & \textcolor{blue}{4.2021} & 6.1795 &  6.5101 & 5.9201 & 6.4777 & 5.8003 & 6.6420   \\
			& IPPR2:   & \textcolor{blue}{6.8575}  & 6.9256 & 6.9256 & 6.9249 & 6.9256 & 6.9256 & 6.9258   \\
			& I\_SI\_01: & 4.1119 & 3.2955 & 3.3333 & 3.2895 & 3.2895 & 3.3518 & \textcolor{blue}{2.5187}  \\
			& Intelligent Room:   & 5.9152 & 	3.3890 & 3.4871 & 3.3546 & 3.5144 & 3.4951 & \textcolor{blue}{3.3934}  \\
			& Intersection : & \textcolor{blue}{2.6911} & 13.9704 & 13.9690 & 13.9752 & 13.9720 & 13.9726 & 13.1152   \\
			& MPEG4\_40:   & 3.9292 & 4.3346 & 5.6052 & 4.2712 & 5.5649 & 5.6711 & \textcolor{blue}{3.7329}  \\
			& PETS2006:   & 6.5818 & \textcolor{blue}{4.7506} & 5.5686 & 4.7573 & 	5.4968 & 5.6115 & 5.5221 \\
			& Fluid Highway:   & \textcolor{blue}{4.3549} & 12.1362 & 9.3360 & 10.1921 & 9.3345 & 9.2739 & 10.1913 \\
			& Highway : & 4.8638 & 4.0454 & 4.1048 & 	\textcolor{blue}{4.0381} & 4.0901 & 4.0941 & 4.0762 \\
			& Skating :   & \textcolor{blue}{5.855}  & 26.0429 & 25.9610 & 26.1047 & 26.0922 & 25.9509 & 25.7092 \\
			& Street Corner at Night: & \textcolor{blue}{9.5308} & 10.2057 & 10.1120 & 10.1791 & 10.0807 & 10.1170 & 12.9509  \\
			& wetSnow :   & \textcolor{blue}{12.3658} & 37.6461 & 37.7272 & 38.1130 & 37.7126 & 37.7054 & 38.6056  \\
			\hline
			Average AGE:       &          & \textcolor{red}{5.3666} & 9.6654 & 12.3337 & 9.5385 & 15.8439 & 12.6639 & 9.7332  \\
			\hline
			Intermittent Motion:         & AVSS2007:   & \textcolor{blue}{7.3008} & 21.3837 & 21.3776 & 21.3957 & 21.3896 & 21.3746 & 35.5689 \\
			& CaVignal :   & 13.9885 & 1.6927 & 1.7240 & 1.7131 & 1.7379 & 1.7182 & \textcolor{blue}{1.3504}\\
			& Candela m1.10:   & 8.6512 & \textcolor{blue}{3.8845} & 3.8889 & 3.9102 & 3.8977 & 3.9043 & 5.4697  \\
			& I\_CA\_01: & 16.8939 & 15.4821 & 15.4496 & 15.4985 & 15.4312 & 15.4297 & \textcolor{blue}{14.6558}  \\
			& I\_CA\_02: & 13.4803 & 9.9255 & 6.6146 & 9.8810 & \textcolor{blue}{6.2029} & 7.1204 & 9.8810  \\
			& I\_MB\_01: & 9.2338 & 8.1882 & 7.3860 & 8.0478 & \textcolor{blue}{7.1827} & 7.6550 & 11.5584  \\
			& I\_MB\_02: & 9.5397 & 8.6324 & 8.6360 & 8.6361 & 8.6307 & 8.6353 & \textcolor{blue}{3.6324} \\
			& Teknomo:   & \textcolor{blue}{4.8436} & 6.7690 & 6.7382 & 6.7388 & 6.7315 & 6.7312 & 6.7310  \\
			& UCF-traffic:   & \textcolor{blue}{4.1126} & 33.0448 &  33.0449 & 33.0464 & 	33.0426 & 33.0432 & 32.9837 \\
			& Uturn:   & \textcolor{blue}{7.4448} & 23.4947 & 23.5190 & 23.4939 & 23.5187 & 23.5163 & 21.2872 \\
			& Bus Station:   & 8.9723 & 3.5451 & \textcolor{blue}{3.5409} & 3.5513 & 3.5525 & 3.5474 & 6.5359 \\
			& Copy Machine:   & 7.3156 & 8.1650 & 8.2640 & 8.1819 & 	8.2836 & 8.2483 & \textcolor{blue}{4.9248}   \\
			& Office:   & 16.6488 & 9.2656 & 9.1716 & 9.2710 & 9.1694 & 9.2024 & \textcolor{blue}{3.3454}  \\
			& Sofa:    & 4.9927 & 4.2697 &  4.2711 & 4.2637 & 4.2708 & 4.2616 & \textcolor{blue}{4.1817} \\
			& Street Corner:   & 8.9535 & \textcolor{blue}{7.6411} & 7.7734 & 7.6425 & 7.8462 & 7.6832 & 27.5613   \\
			& Tramstop:   & 7.1293 & 2.4173 & 2.4268 & 2.4282 & 	2.4483 & 2.4153 & \textcolor{blue}{2.4079} \\
			\hline
			Average AGE:    &          & \textcolor{red}{9.3438} & 10.4876 & 10.2392 & 10.4812 & 10.2085 & 10.2804 & 12.0047  \\
			\hline
			Jitter:           & CMU:     & 8.1714 & 7.3476 & 6.9292 & 7.3197 & 7.7878 & 7.6034 & \textcolor{blue}{6.8975} \\
			& I\_MC\_02: & \textcolor{blue}{9.0549} & 15.7418 & 13.9334 & 15.4235 & 15.4017 & 15.6302 & 15.9440 \\
			& I\_SM\_04: & 4.5583 & 3.3464 & \textcolor{blue}{2.5355} & 3.0923 & 3.7768 & 4.3339 & 4.1406   \\
			& O\_MC\_02: & 12.6371 & 16.3119 & 17.3914  & 16.6375 & 	16.0443 & 16.4781 & \textcolor{blue}{12.3657}\\
			& O\_SM\_04: & \textcolor{blue}{7.7459} & 12.0224 & 12.0262 & 13.2998 & 15.6505 & 13.9053 & 15.6806\\
			& Badminton:   & 14.2284 & 16.9398 & 17.1787 & 16.4044 & 	16.6059 & 14.2486 & \textcolor{blue}{6.6003} \\
			& Boulevard:   & \textcolor{blue}{11.5450} & 19.4259 & 15.4555 & 16.6356 & 	20.0932 & 16.9604 & 23.8209\\
			& Side Walk:   & \textcolor{blue}{14.9378} & 24.7621 & 24.1964 & 22.8313 &  16.5027 & 15.8949 & 18.4447 \\
			& Traffic:   & 21.3232 & 7.5524 & 24.5624 & 8.6431 & 	7.5449 & \textcolor{blue}{6.7522} & 26.5434  \\
			\hline
			Average AGE:      &          & \textcolor{red}{11.5780} & 13.7167 & 14.9121 & 13.3652 & 13.2675 & 12.4230 & 14.4931 \\
			\hline
			Very Short:       & CUHK Square:  & \textcolor{blue}{2.8429} & 5.4994 & 4.8949 & 5.8176 &  5.2220 & 5.0429 & 6.2694  \\
			& Dynamic Background :   & 13.7524 & 7.7233 & 7.8492 & 7.5747 & 7.9276 & 7.3880 & \textcolor{blue}{7.3760} \\
			& MIT:     & \textcolor{blue}{3.5838} & 4.9527 & 5.7849 & 4.4991 & 5.8378 & 5.2764 & 4.9524 \\
			& Noisy Night :   & \textcolor{blue}{3.9116} & 6.1301 & 5.5040 & 6.3509 & 5.3378 & 5.6906 & 5.4483   \\
			& Toscana:   & 11.5422 & 8.7331 & 6.4773 & 7.4022 & 6.8142 & \textcolor{blue}{6.3869} & 7.4014   \\
			& Town Center : & 4.1427 & 4.4226 & 4.4247 & 4.2329 &  \textcolor{blue}{3.8596} & 3.9657 & 4.4225  \\
			& Two Leave Shop1cor:   & 10.0183 & 4.0515 & 4.0172 & 4.2124 & \textcolor{blue}{3.9300} & 3.8685 & 4.0503  \\
			& Pedestrians:   & 5.0736 & 5.0318 & \textcolor{blue}{4.9441} & 4.9996 & 4.9974 & 4.9682 & 5.0225 \\
			& People In Shade:   & 6.9680 & 9.0900 & 6.5455 & 10.7783 & \textcolor{blue}{3.6842} & 9.3889 & 10.7812   \\
			& SnowFall:   & \textcolor{blue}{5.2768} & 32.8871 & 31.0542 & 31.2511 & 31.8320 & 	30.3902 & 34.2603 \\
			\hline
			Average AGE:    &          & \textcolor{red}{6.7112} & 8.8522 & 8.1496 & 8.7119 & 7.9443 & 8.2366 & 8.9984 \\
			\hline
			Illumination Changes: & Camera Parameter:   & 6.2206 & 75.1204 & 6.1471 & \textcolor{blue}{6.1126} &  6.1389 & 6.1475 & 45.2837 \\
			& Dataset3 Camera1 :   & 14.5708 & 23.3046 & 22.0816 & 22.5116 & 22.0816 & 22.0816 & \textcolor{blue}{2.8850}  \\
			& Dataset3 Camera2:   & 18.7047 & 6.5041 & 5.7156 & 5.8965 &  5.7156 & 5.7156 & \textcolor{blue}{3.7555} \\
			& I\_IL\_01: & \textcolor{blue}{7.4329} & 8.3048 & 23.6585 & 23.5775 & 23.6585 & 23.6585 & 22.4594   \\
			& I\_IL\_02: & 19.3833 & 8.4842 & 7.5423 & 7.4007 &  7.5423 & 7.5423 & \textcolor{blue}{5.1225}\\
			& Cubicle:   & \textcolor{blue}{11.4636} & 26.1490 & 19.4842 & 31.2116 & 19.4842 & 19.4842 & 13.0519 \\
			\hline
			Average AGE:    &          & \textcolor{red}{12.9627} & 24.6445 & 14.1049 & 16.1184 & 14.1035 & 14.1049 & 15.4263 \\
			\hline
			Very Long:        & Bus Stop Morning :   & \textcolor{blue}{3.1641} & 5.6652 & 5.7055 & 5.6396 & 5.6739 & 5.6794 & 5.7419 \\
			& Dataset4 Camera1:   & 6.7405 & 3.1857 & 3.1886 & 3.1876 & 3.1794 & 3.1948 & \textcolor{blue}{3.1661}   \\
			& Ped And Storrow Drive:   & 8.5110 & 5.5780 & 5.0913 & 5.4323 & 5.3057 & 5.2445 & \textcolor{blue}{4.5065} \\
			& Ped And Storrow Drive3 :   & \textcolor{blue}{2.8661} & 3.5503 & 3.6693 & 3.5531 & 3.6100 & 3.5598 & 3.9688 \\
			& Terrace :   & \textcolor{blue}{6.0016} & 19.9480 & 18.9514 & 19.1109 & 19.0254 & 19.0258 & 10.2339 \\
			\hline
			Average AGE:    &          & \textcolor{blue}{5.4567} & 7.5854 & 7.3212 & 7.3847 & 7.3589 & 7.3409 & 5.5234 \\ \hline
			\textcolor{red}{Average AGE of all categories}:            &          &
			\textcolor{red}{8.7237} & 13.2359 & 11.9362 & 11.9229 & 12.1183 & 11.5934 & 11.6340 \\ \hline
		\end{tabular}
	}
\end{table*}
\begin{figure*}[t]
	\centering
	\includegraphics[scale=0.4]{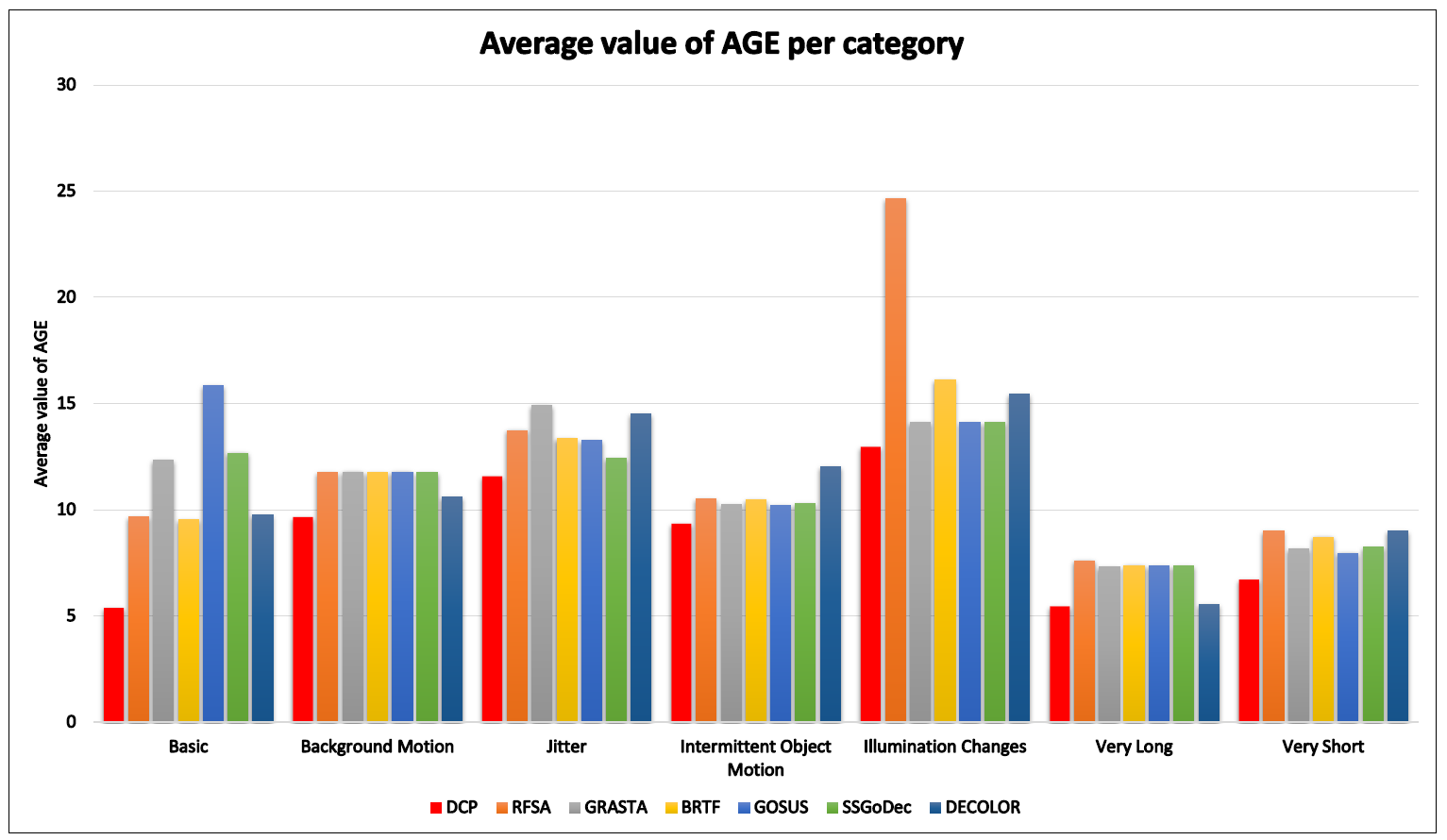}
	\caption{Performance comparison of each method on the basis of AGE according to each category on SBM.net dataset.}
	\label{fig_AGEcate}
\end{figure*}

\textbf{Category: Basic} contains $16$ video sequences (Table \ref{all_dataset}). In almost all video sequences our proposed approach DCP performed well. DCP achieved an average AGE of $5.367$ (visual results in Figure \ref{fig_AGEcate}) among all the compared methods because this category contains relatively simple scenes for background estimation. It can be seen in the Table \ref{all_dataset}, that RFSA, BRTF and DECOLOR almost achieved equal and second lowest score of AGE but GOSUS and GRASTA achieved a bit bit higher values of AGE. GOSUS suffered performance degradation among all compared methods. In terms of qualitative analysis DCP estimated better background as compared to all the methods, results are shown in the Figure \ref{fig_vis}, (c), (d) and (e). The reason is that the context for video sequences of `Wet-snow', `Skating' and `Street Corner at Night'  is homogeneous in the whole frame as background pixel values. This key aspect is favorable for our proposed method.

\textbf{Category: Intermittent Motion}  contains $16$ video sequences (Table \ref{all_dataset}). This category has video sequences which contain ghosting artifacts in the detected motion. DCP performed well in this category  by achieving lowest AGE score of $9.344$ among all compared techniques. Methods including RFSA, GRASTA, BRTF, GOSUS and SSGoDec achieved almost equal and  higher score of AGE (Figure \ref{fig_AGEcate} and Table \ref{all_dataset}).  DECOLOR has the highest error rate in background estimation for this category. The ghosting artifacts pose big challenge for all algorithms as  the foreground becomes the part of background, resulting in failure of accurate background recovery model.

\textbf{Category: Jitter} contains $9$ video sequences (Table \ref{all_dataset}). DCP achieved lowest average gray level error among all the compared methods due to the fact that camera jitter contains videos sequences with blurry context and such context is easy to predict by our proposed method. RFSA, BRTF and GOSUS achieved higher  AGE score in this category while GRASTA and DECOLOR showed performance degradation among all compared methods. It can be seen in  Figure \ref{fig_intro} (e) that GRASTA was not able to recover clean background while DCP estimated it accurately. SSGoDec is also able to recover clean background as shown in  Figure \ref{fig_vis} (h) with low AGE score.

\textbf{Category: Very Short} contains $10$ video sequences each having only few frames (Table \ref{all_dataset}). DCP achieved the lowest AGE score in this category too. GOSUS also performed well and achieved the second lowest AGE score as shown in  Table \ref{all_dataset}. However RFSA, GRASTA, BRTF, DECOLOR and SSGoDec achieved almost equal score of AGE  among all the compared methods. In terms of qualitative analysis, it can be seen in  Figure \ref{fig_vis} (c) that for instance, the video sequence `SnowFall', DCP achieved the lowest score of AGE. It is due to the fact that in case of bad weather snow or rain the context of the videos gets blurry which is rather easy for DCP to estimate.

\textbf{Category: Illumination Changes} contains $6$ video sequences (Table \ref{all_dataset}). This category pose a great deal of challenge for all the methods. DCP managed to get lowest AGE score among all the compared methods due to the fact that context prediction in low light and with less sharp details is rather favorable condition for our proposed method.  GRASTA, GOSUS and SSGoDec also performed well and achieved second lowest AGE score among all compared methods. BRTF and DECOLOR almost get equal AGE score. RFSA has the highest error rate as shown in Table \ref{all_dataset}, because of the spatiao-temporal smoothness of foreground, and the correlation of background constraint.

\textbf{Category: Very Long} contains $5$ video sequences containing thousands of frames (Table \ref{all_dataset}). Among all the compared methods only DCP and DECOLOR performed well with the lowest AGE score of $5.457$ and $5.524$ respectively. However all methods except DCP and DECOLOR achieved nearly equal score of average gray level error for background estimation (Table \ref{all_dataset}). For instance in the case of DCP, video sequence "Bus Stop Morning" achieved the lowest AGE score of $3.164$ among all compared methods, its visual result is shown in Figure \ref{fig_intro} (d).
\subsubsection{Overall Performance Comparison of DCP for Background Estimation}
Upon averaging the results from all the 7 categories, DCP achieved an average gray level error to be $8.724$ which is minimum among all the compared methods as shown in Figure \ref{fig_metric} (a). For fair comparison and evaluation other than AGE, results of $5$ other metrics have also been calculated.
In Figure \ref{fig_metric} (b), pCEPS which is Percentage of Clustered Error Pixels is minimum for DCP among all compared methods.  BRTF, GOSUS and SSGoDec has  higher value than DCP. The other three methods GRASTA, RFSA and DECOLOR achieved almost equal and highest score of pCEPs. The metric pEPs which is basically Percentage of Error Pixels, is aimed to get minimum score for accurate background  estimation (Figure \ref{fig_metric} (c)). Among all the compared methods only DCP achieved the minimum score while all compared methods showed minimal difference in their pEPs score. In Figure
\ref{fig_metric} (d) CQM: (Color image Quality Measure), DCP achieved the maximum (best) score for this metric. It can also be seen in the visual results ( Figure \ref{fig_vis} (d), (e), (f), (g), (h) and (i)) that color quality of some background images extracted by compared methods, is different from input images, ground truths and backgrounds estimated by DCP. Due to this reason all compared methods have different scores of CQM metric. In Figure \ref{fig_metric} (e), PSNR: (Peak-Signal-to-Noise-Ratio)
and Figure \ref{fig_metric} (f) MSSSIM: (MultiScale Structural Similarity Index) should have a highest value for best performance and DCP achieved it efficiently. The proposed DCP algorithm achieved best scores in all mentioned metrics, as compared to the $6$ methods. 

\begin{table*}\label{cdnet_table}
	\centering
	\caption{Comparison of $6$ state of the art methods with the proposed DCP algorithm by using $F$ measure on CDnet2014 dataset. The first highest and the second highest scores for each category is shown in red and blue color respectively.}
	\scalebox{0.75}{
		\label{f_scoretable}
		\begin{tabular}{l*{6}{c}r}
			\hline
			Categories  & MSSTBM \cite{lu2014multiscale} & GMM-Zivkovic \cite{zivkovic2004improved} & CP3-Online \cite{liang2015co} & GMM-Stauffer \cite{stauffer1999adaptive} & KDE-ElGammal \cite{elgammal2000non} & RMOG \cite{varadarajan2013spatial} & DCP \\
			\hline
			Baseline                    & 0.8450 & 0.8382 & \textcolor{blue}{0.8856} & 0.8245 & \textcolor{red}{0.9092} & 	0.7848 & 0.8187  \\
			Camera Jitter               & 0.5073 & 0.5670 & 0.5207 & 0.5969 &  0.5720 & \textcolor{blue}{0.7010} & \textcolor{red}{0.8376}  \\
			Shadow                      & \textcolor{red}{0.8130} & 0.7232 & 0.6539 & 0.7156 &  0.7660 & \textcolor{blue}{0.8073} & 0.7665   \\
			Dynamic Background          & 0.5953 & 0.6328 & 0.6111 & 0.6330 &  0.5961 & \textcolor{blue}{0.7352} & \textcolor{red}{0.7757} \\
			Thermal                     & 0.5103 & 0.6548 & \textcolor{blue}{0.7917} & 0.6621 &  0.7423 & 0.4788 & \textcolor{red}{0.8212}  \\
			Intermittent Object Motion  & 0.4497 & 0.5325 & \textcolor{red}{0.6177} & 0.5207 &  0.4088 & 0.5431 & \textcolor{blue}{0.5979}  \\
			Bad Weather                 & 0.6371 & 0.7406 & 0.7485 & 0.7380 &  \textcolor{blue}{0.7571} & 0.6826 & \textcolor{red}{0.8212}   \\
			\hline
			Average  &  0.6225 & 0.6736 & \textcolor{blue}{0.7010} & 0.6771 & 0.6833 & 0.6761 &  \textcolor{red}{0.7620}  \\
			\hline
		\end{tabular}
	}
\end{table*}

\subsection{Evaluation of Deep Context Prediction (DCP) for Foreground Detection} \label{cdnet}
We have selected $7$ categories from CDnet2014 \cite{wang2014cdnet} dataset. The results are compared with $6$ state-of-the-art methods, including MSSTBM \cite{lu2014multiscale}, GMM-Zivkovic \cite{zivkovic2004improved}, CP3-Online \cite{liang2015co}, 
GMM-Stauffer \cite{stauffer1999adaptive}, KDE-Elgammal \cite{elgammal2000non} and RMoG \cite{varadarajan2013spatial} by using implementations of the original authors. Foreground detection is compared using Average $F$ measure across all the video sequences within each category. The metrics to calculate $F$ measure are as follows :
\begin{equation} \label{re}
Re =  \frac{T_{p}}{T_{p} + F_{n}},
\end{equation}
\begin{equation} \label{sp}
Sp =  \frac{T_{n}}{T_{n} + F_{p}},
\end{equation}
\begin{equation} \label{fnr}
FNR =  \frac{F_{n}}{T_{p} + F_{n}},
\end{equation}
\begin{equation} \label{pwc}
PWC =  100 \times ( \frac{F_{n} + F_{p}}{T_{p} + F_{n} + F_{p} + T_{n} }),
\end{equation}
\begin{equation} \label{pre}
Pre =  \frac{T_{p}}{T_{p} + F_{p}},~~and
\end{equation}
\begin{equation}
F =  \frac{2 (Pre\times Re)}{Pre + Re},
\end{equation}
where $T_{p}$ is True positives, $T_{n}$ is True negatives, $F_{p}$ is False positives, $F_{n}$ is False negatives, $Re$ is Recall, $Sp$ is Specificity, $FNR$ is False Negative Rate, $PWC$ Percentage of Wrong Classifications, $Pre$ is Precision and $F$ is F-Measure.
Following is the detailed explanation of results on 7 categories of CDnet2014  dataset.

\begin{figure*}[t]
	\centering
	\includegraphics[scale=0.4]{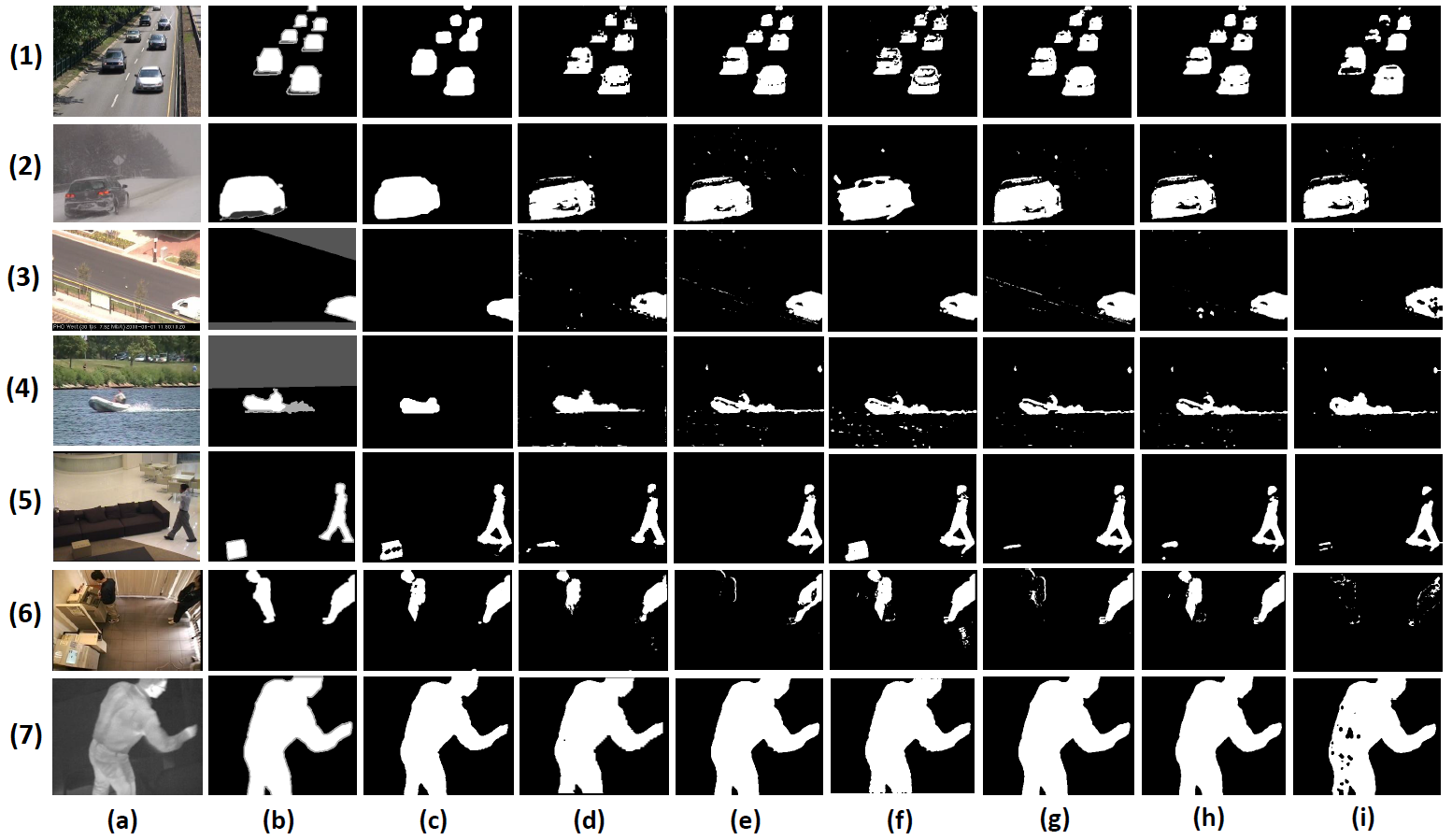}
	\caption{Qualitative results of the proposed DCP method: (a) Seven images from the input video sequences of Cdnet2014 dataset, (b) Ground truth,  (c) Foreground detected by the proposed DCP method, (d) MSSTBM, (e) GMM-Zivkovic, (f) CP3-Online, (g) GMM-Stauffer, (h) KDE-ElGammal, (i) RMOG. From top to bottom: each input sequence is selected from different category: (1) sequence `Highway' from `Baseline', (2) `Snowfall' from `Bad Weather', (3) `Boulevard' from `Camera Jitter', (4) `Boats' from `Dynamic Background', (5) `Sofa' from `Intermittent Object Motion', (6) `Copy Machine' from `Shadow', and (7) `Library' from `Thermal'.}
	\label{fig_vis_cdnet}
\end{figure*}

\textbf{Category: Baseline} in CDnet2014 dataset contains $4$ video sequences. The average F measure score across all $4$ video sequences is shown in  Table \ref{f_scoretable}. All the compared methods including DCP achived more than $0.8$ score for this category (Table \ref{f_scoretable}). However KDE-ElGammal successfully got the highest score leaving CP3-Online on second position among all compared methods. Although DCP achieved more than $0.8$ F measure score but still it was not able to beat KDE-ElGammal method due to the fact that successive opening and closing on noisy video frames lead to false detection. Visual results are shown in  Figure \ref{fig_vis_cdnet} first row.

\textbf{Category: Camera Jitter} also contains $4$ video sequences.   DCP achieved the highest  F measure among all the compared methods, as shown in Table \ref{f_scoretable}. It is due to the fact that blurry context because of  camera jitter is easy to predict by our proposed method for accurate background estimation. Afterwards the binarized thresholded difference of the estimated background and current frame erodes the noisy pixels of background in successive opening and closing operations. This leads us to get accurate foreground detection with less missing pixel values of foreground objects (Figure \ref{fig_vis_cdnet}: $3^{rd}$ row). RMOG also performed well in this category and achieved the second best score among all compared methods.

\textbf{Category: Shadow} contains $6$ video sequences. MSSTBM achieved the highest score among all compared methods with RMOG as second best score.  This category posed  challenge to our proposed method as sometimes shadows got replicated in the context prediction algorithm which  generates errors in background estimation as well as foreground detection. In our proposed method the opening and closing of the binarized thresholded difference frame successfully filled the missing values in the foreground detection as shown in the Figure \ref{fig_vis_cdnet}: $6^{th}$ row as compared to all methods. This leads DCP to achieved $3^{rd}$ best F measure in this category. 

\textbf{Category: Dynamic Background} also contains $6$ video sequences.   DCP achieved the highest averaged F measure among all the compared methods, see Table \ref{f_scoretable}. The homogeneous context in video sequences of this category is a favorable condition for our proposed method. RMOG also performed well and achieved the second best  F measure score. The qualitative results are as shown in Figure \ref{fig_vis_cdnet}. It can be seen in the visual results that successive opening and closing with a suitable SE removed the noisy pixel values of moving background.

\textbf{Category: Thermal} contains $5$ video sequences that have been captured by far-infrared camera. DCP  achieved highest averaged F measure score among all compared methods, while CP3-Online is the second best. It is because of the same reason as explained in previous category. The homogeneous context is one of the major key for accurate background estimation of DCP, and it leads to noise-less foreground detection.  Figure \ref{fig_vis_cdnet}: $7^{th}$ row shows that all methods including DCP accurately detected foreground object except RMOG which contains missing pixel values within detected foreground object.

\textbf{Category: Intermittent Object Motion} contains $6$ video sequences with scenarios known for causing “ghosting” artifacts in the detected motion, i.e., objects move, then stop for a short while, after which they start moving again. DCP  achieved the highest average F measure score in this category, while RMOG is the second best among all the compared methods. The main reason behind this is, our proposed approach does not contain any motion-based constraints for moving foreground objects. Since all the compared methods contain constraints on the motion of the foreground objects, which if violated lead to false detection and low F measure score. The visual results in  Figure \ref{fig_vis_cdnet}: $5^{th}$ row show that the foreground objects vanish if motion-based constraints are violated.\\
\textbf{Category: Bad Weather} contains $4$ video sequences captured in challenging winter weather conditions, i.e., snow storm, snow on the ground, and fog. DCP achieved highest averaged F measure among all compared methods while  KDE-ELGammal is the second best method. This category is another example of homogeneous context in video sequences. It can be seen in the visual results, Figure \ref{fig_vis_cdnet}: $2^{nd}$ row, DCP estimated the almost accurate foreground object with no unconnected noisy pixels of background as compared to  the other methods.
\subsubsection{Overall Performance Comparison of DCP for Foreground Detection}
Table \ref{f_scoretable} shows that  DCP achieved the highest average  F measure score over all categories.  CP3-Online is the $2^{nd}$ best algorithm. GMM-Stauffer, GMM-Zivkovic, KDE-ElGammal and RMOG  achieved almost equal F measure with a minimal difference. MSSTBM achieved the lowest score among all the compared methods (Table \ref{f_scoretable}).
For better foreground detection the aim of the metrics (defined in \eqref{re}, \eqref{sp}, \eqref{fnr}, \eqref{pwc}, and \eqref{pre}) is to maximize the values of Re, Sp and Precision and minimize the values of FNR and PWC. The proposed DCP algorithm achieved top score in Re and FNR which is $0.809$ and $0.191$ respectively among all the compared methods. It means that  more correct  detection and less incorrect detection of foreground objects  by our proposed method. Moreover for metrics like PWC, Sp and Precision, DCP achieved $2.671$ , $0.977$ and $0.773$ best scores respectively which are higher than most of the methods.
\begin{figure}[t]
	\centering
	\includegraphics[scale=0.3]{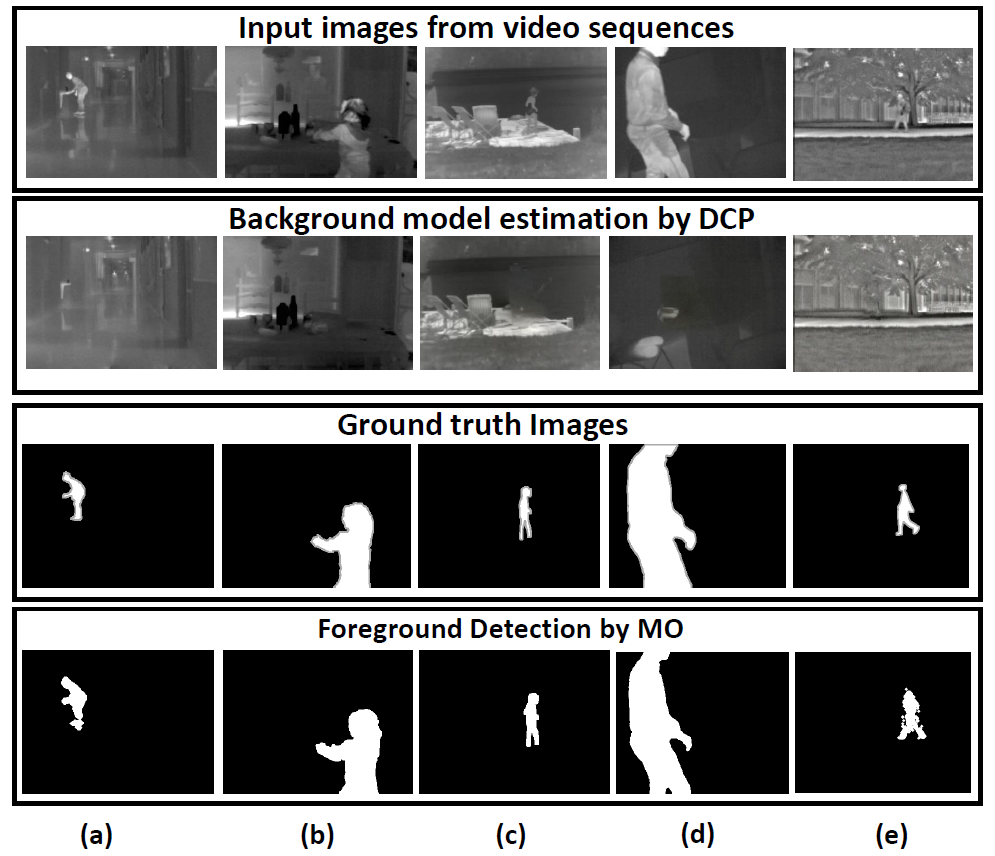}
	\caption{Estimated background examples  from the CDnet2014 dataset : all sequences are from the category "Thermal". (a) Sequence "Corridor". (b) Sequence "Dining Room". (c) Sequence "Lake Side". (d) Sequence "Library". (e) Sequence "Park". In all of these video sequences, DCP estimated an accurate the background which leads to better foreground detection as well.}
	\label{cdnet_thermal}
\end{figure}
\subsection{Performance of DCP on the basis of Homogeneous Context} \label{1_cd}
As explained in Section \ref{Proposed},  our proposed method estimates the  background on the basis of context prediction, so in this section we discuss the key aspects of DCP on the type of contexts present in the video sequences containing different scenes specifically for the application of background estimation.

\begin{figure}[t]
	\centering
	\includegraphics[scale=0.34]{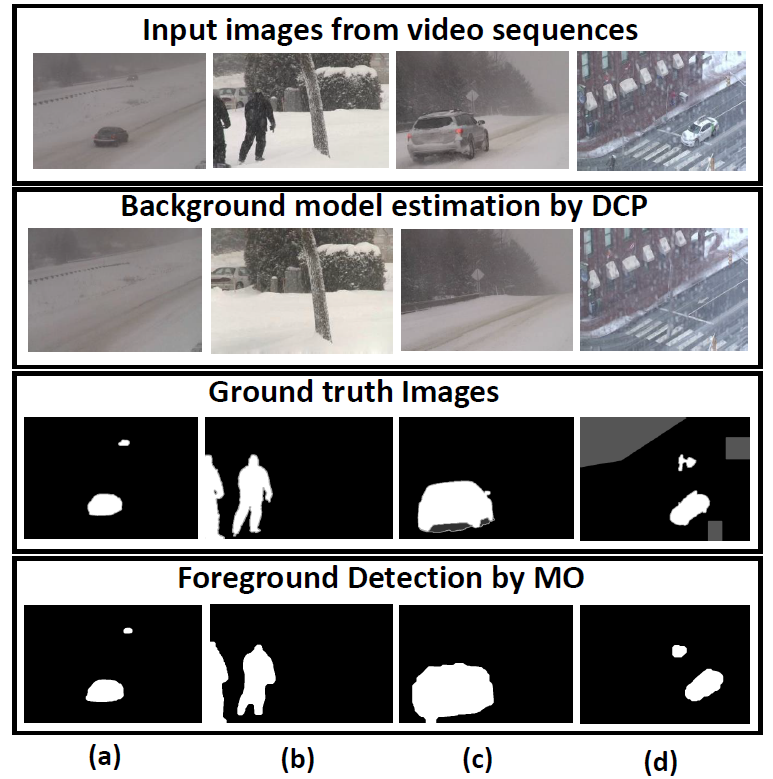}
	\caption{Estimated background examples  from the CDnet2014 dataset: sequences in (a) are from the category `Bad Weather' named `Blizzard'. (b) Sequence `Skating'. (c) Sequence `SnowFall'. (d) Sequence `WetSnow'. In all of these video sequences, DCP estimated an accurate background which leads to better foreground detection as well.}
	\label{cdnet_BW}
\end{figure}

Table \ref{all_dataset} shows that for all categories,  AGE score is different even for individual videos per category for all the compared methods including DCP. The  reason behind that is, the context of every video is different with different kinds of indoor outdoor scenes. Therefore, for  compared methods including DCP the average gray level score is different and quite challenging in some cases as well. For convenience, we are targeting the discussion of homogeneous context to few video sequences in SBM.net  and CDnet2014 dataset.  We have selected $2$ categories from CDnet2014 dataset on the basis of their homogeneous context in the video sequences. Category wise discussion is as follows:

\textbf{Category: Bad Weather} is a similar context example from CDnet2014 dataset.  Figure \ref{cdnet_BW} shows the visual result of video sequence "blizzard" however other three video sequences are same, "skating", "wetsnow" and "snowfall" from category "Basic" in SBM.net dataset. These video sequences have minimum score of AGE  and their visual result are shown in Figure \ref{fig_vis} (c): $2^{nd}, 4^{th}$ and $6^{th}$ row and Figure \ref{cdnet_BW}.

\textbf{Category:Thermal} is another challenging category in CDnet2014 which includes videos that have been captured by far-infrared cameras. The interesting fact about this category is it includes video sequences with thermal artifacts such as heat stamps, heat reflection on floors, windows, camouflage effects, and a moving object may have the same temperature as the surrounding regions \footnote{\url{http://jacarini.dinf.usherbrooke.ca/datasetOverview/}}. It is very favorable environment for DCP for context prediction. The visual results of all $5$ video sequences for this category are  shown in Figure \ref{cdnet_thermal}.

\subsection{Failure Cases for DCP}
\begin{figure}
	\centering
	\includegraphics[scale=0.45]{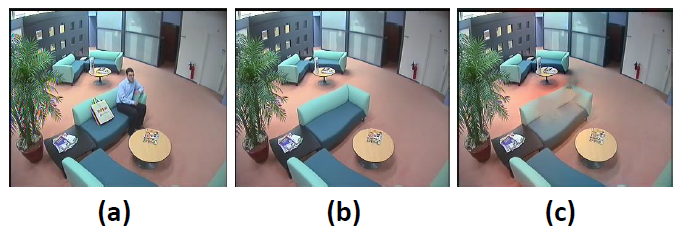} 
	\caption{Estimated background example  from the SBM.net dataset: (a) Sequence $Candela~m1.10$ (b) Ground truth (c) Estimated background by DCP. Table \ref{all_dataset} shows that for category `Intermittent Motion'  AGE of DCP is maximum than all the compared methods.}
	\label{fig_failure}
\end{figure}

Although DCP achieved good performance in most of the cases, still it has some limitations and failure cases. Estimation of complex background structures (Figure \ref{fig_failure}) and large scale foreground objects is quite challenging. The limitation of the proposed method involves large sized foreground objects to be accurately inpainted. In these cases,  the network is not able to  properly fill the region in an irregular shape. We used Poisson blending technique to transform center region inpainting context to irregular region one.

\section{Conclusion} \label{con}
In this work  a unified method  `Deep Context Prediction' (DCP) is proposed for background estimation and foreground segmentation using GAN and image inpainting. The proposed  method is based on an unsupervised visual feature learning based hybrid GAN  for context prediction along with semantic inpainting network for texture optimization.  Solution of random region inpainting is also proposed by using  center region inpainting and  Poisson blending. The proposed DCP algorithm is compared with six existing algorithms for background estimation on SBM.net dataset. The proposed  algorithm has outperformed these compared methods with a significant margin. The proposed  algorithm is also compared with six foreground segmentation methods on CDnet2014 dataset. On the average, the proposed algorithm has outperformed these algorithms. These experiments demonstrate the effectiveness of the proposed approach compared to the existing algorithms. The proposed algorithm has demonstrated excellent results in bad weather and thermal imaging categories in which most of the existing algorithms suffer from performance degradation.


\begin{acknowledgements}
This study was supported by the BK21 Plus project (SW Human Resource Development Program for Supporting Smart Life)
funded by the Ministry of Education, School of Computer Science and Engineering, Kyungpook National University, Korea (21A20131600005).
\end{acknowledgements}

\bibliographystyle{spbasic}      
\bibliography{mybibfile} 

\vspace{2ex}\noindent{\bf Maryam Sultana} is a PhD student at Virtual Reality Lab, School of Computer Science and Engineering, Kyungpook National University Republic of Korea. She received her M.Sc. and M.Phil. degrees in electronics from Quaid-i-Azam university Pakistan in 2013 and 2016, respectively. Her research interests include background modeling, foreground object detection and generative adversarial networks.\\

\noindent {\bf Arif Mahmood} received the master’s and Ph.D.
degrees in computer science from the Lahore University of Management Sciences, Lahore, Pakistan, in 2003 and 2011, respectively. He was a Research Assistant Professor with the School of Computer Science and Software Engineering, The University of Western Australia (UWA), where he was involved in hyper-spectral object recognition and action recognition using depth images. He was a
Research Assistant Professor with the School of Mathematics and Statistics, UWA, where he was involved in the characterizing structure of complex networks using sparse subspace clustering. He is currently a Post-Doctoral Researcher with the Department of Computer Science and Engineering, Qatar University, Doha. He has performed research in data clustering, classification, action, and object recognition. His major research interests are in computer vision and pattern recognition, action detection and person segmentation in crowded environments, and background-foreground modeling in complex scenes.\\

\noindent {\bf Sajid Javed} is currently a Post-doctoral research fellow in the Department of Computer Science, University of Warwick, United Kingdom. He obtained his Bs.c (hons) degree in Computer Science from University of Hertfordshire, UK, in 2010. He joined the Virtual Reality Laboratory of Kyungpook National University, Republic of Korea, in 2012 where he completed his combined Master's and Doctoral degrees in Computer Science. His research interests include background modeling and foreground object detection, robust principal component analysis, matrix completion, and subspace clustering.\\

\noindent{\bf Soon Ki Jung} is a professor in the School of Computer Science and Engineering at Kyungpook National University, Republic of Korea. He received his MS and PhD degrees in computer science from Korea Advanced Institute of Science and Technology (KAIST), Korea, in 1992 and 1997, respectively. He has been a visiting professor at University of Southern California, USA, in 2009. He has been an active executive board member of Human Computer Interaction, Computer Graphics, and Multimedia societies in Korea. Since 2007, he has also served as executive board member of IDIS Inc. His research areas include a broad range of computer vision, computer graphics, and virtual reality topics.


%
%

\end{document}